\documentclass[10pt,twocolumn,letterpaper]{article}

\usepackage{cvpr}
\usepackage{times}
\usepackage{epsfig}
\usepackage{graphicx}
\usepackage{amsmath}
\usepackage{bbm}
\usepackage{amssymb}
\usepackage{algorithm}
\usepackage{algpseudocode} 
\usepackage{booktabs}
\usepackage{comment}
\usepackage{capt-of,etoolbox}
\usepackage{tabularx}
\usepackage{xspace}
\usepackage{kotex}
\usepackage{subcaption}
\usepackage{subfiles}
\usepackage{setspace}
\usepackage[toc,page]{appendix}
\usepackage{mathptmx}

\usepackage{multirow, makecell}
\usepackage{colortbl}
\definecolor{Gray}{gray}{0.9}

\DeclareMathOperator*{\argmax}{\arg\!\max}
\usepackage{xcolor}

\newif\ifdraft\drafttrue
\ifdraft

\newcommand\yj[1]{\textcolor{black}{#1}} 
\newcommand\jw[1]{\textcolor{black}{{#1}}} 
\else

\newcommand\yj[1]{#1}

\newcommand\jw[1]{#1}

\fi

\newcommand{\memname}{RM\xspace}
\newcommand{\our}{RM\xspace}
\newcommand{\fullour}{Rainbow Memory\xspace}

\newcommand\blfootnote[1]{%
  \begingroup
  \renewcommand\thefootnote{}\footnote{#1}%
  \addtocounter{footnote}{-1}%
  \endgroup
}

\usepackage[pagebackref=true,breaklinks=true,colorlinks,bookmarks=false]{hyperref}

\cvprfinalcopy 

\def\cvprPaperID{4189} 

\ifcvprfinal\fi

\begin{document}
\makeatletter

\makeatother

\title{Rainbow Memory: Continual Learning with a Memory of Diverse Samples}

\author{Jihwan Bang$^{1,*}$\hspace{1.5em}Heesu Kim$^{2,3,*}$\hspace{1.5em}YoungJoon Yoo$^{2,3}$\hspace{1.5em}Jung-Woo Ha$^{2,3}$\hspace{1.5em}Jonghyun Choi$^{4,\dagger}$\\
{Search Solutions, Inc$^1$\hspace{3em}NAVER CLOVA$^2$\hspace{3em}NAVER AI Lab$^3$\hspace{3em}GIST$^4$}\\
{\tt\small {\{jihwan.bang,heesu.kim89,youngjoon.yoo,jungwoo.ha\}@navercorp.com, jhc@gist.ac.kr}}
}

\maketitle

\blfootnote{\hspace{-1.8em}$^*$ indicates equal contribution. $^\dagger$ indicates corresponding author.}

\begin{abstract}
Continual learning is a realistic learning scenario for AI models. 
Prevalent scenario of continual learning, however, assumes disjoint sets of classes as tasks and is less realistic, rather artificial.
Instead, we focus on `blurry' task boundary; where tasks shares classes and is more realistic and practical.
To address such task, we argue the importance of diversity of samples in an episodic memory. 
To enhance the sample diversity in the memory, we propose a novel memory management strategy based on per-sample classification uncertainty and data augmentation, named \fullour (\memname).
With extensive empirical validations on MNIST, CIFAR10, CIFAR100 and ImageNet datasets, we show that the proposed method significantly improves the accuracy in blurry continual learning setups, outperforming state of the arts by large margins despite its simplicity. 
Code and data splits will be available in \small\url{https://github.com/clovaai/rainbow-memory}.
\end{abstract}

\section{Introduction}
\label{sec:introduction}

\jw{Continual learning (CL) or class incremental learning (CIL) is known to particularly suffer from the catastrophic forgetting with respect to model generalization, due to inaccessibility to the data of previous tasks. 
The challenge lies in the continuously changing class distributions of each task given a task stream. 
Most AI models suffer from such real-world application scenarios across domains~\cite{Prabhu2020GDumbAS, houlsby2019parameter,li2019compositional}. 
To address the issue of changing data distribution for continual learning, there are many proposals in the literature, such as momentum matching~\cite{Lee2017OvercomingCF}, sample generation~\cite{Shin2017ContinualLW, Wu2018MemoryRG, aydoreDF18, ven2018three}, regularization on parameters~\cite{ewc, rwalk}, and sampling-based memory management~\cite{Prabhu2020GDumbAS, icarl}.} 

\begin{figure}[t!]
    \centering
    \includegraphics[width=0.99\columnwidth]{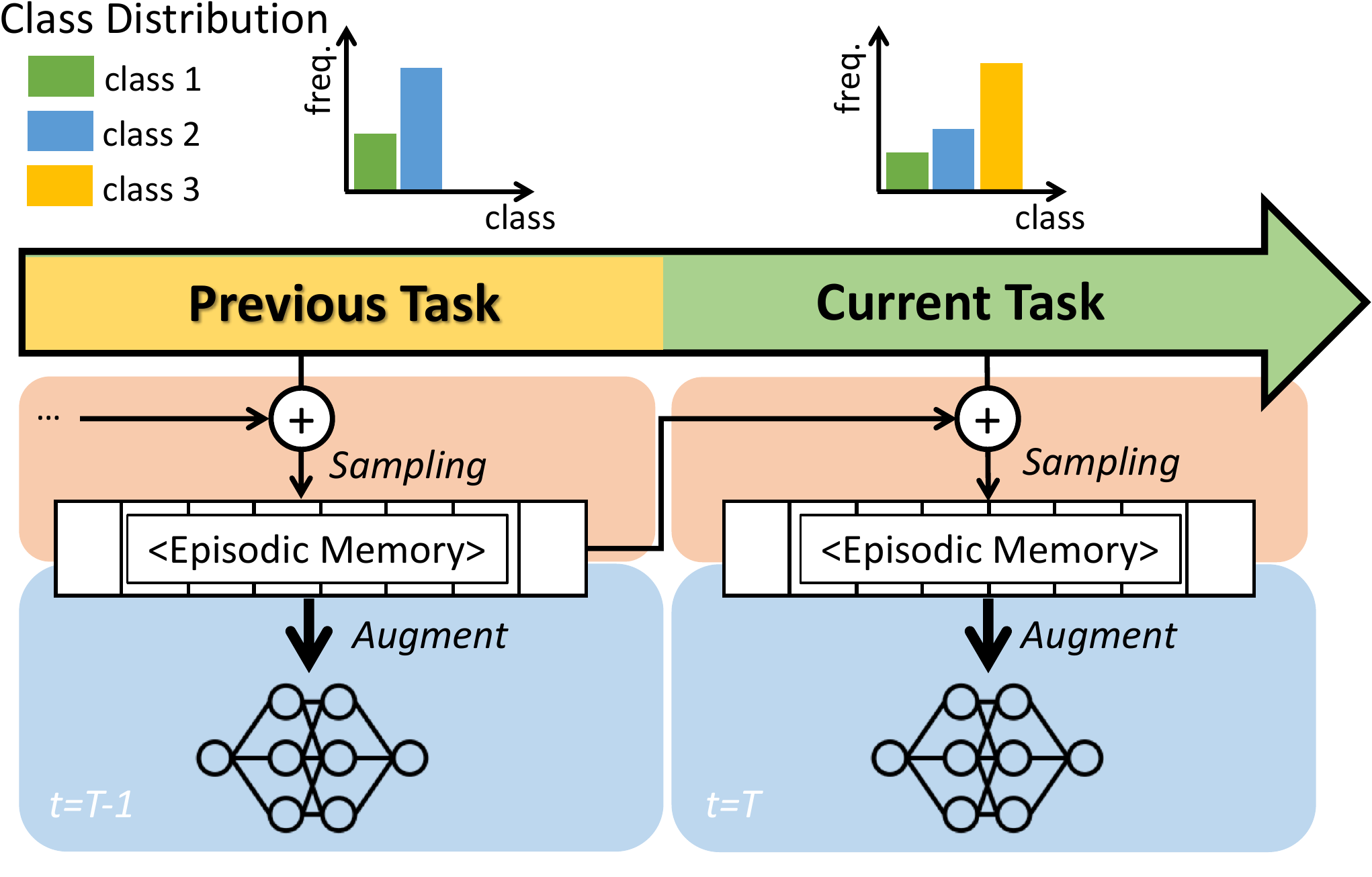}
    \caption{Blurry-CIL (class incremental learning) setup (top) and overview of our proposed approach (bottom). In the blurry-CIL, the tasks share classes, contrary to conventional disjoint-CIL. Proposed memory management strategy updates an episodic memory with samples of the current task to keep diverse exemplars in the memory. Data augmentation (DA) further enhances the diversity of the exemplars in the memory.}
    \vspace{-1em}
    \label{fig:blurry-CIL}
\end{figure}

\begin{figure}[t!]
    \centering
    \includegraphics[width=1\columnwidth]{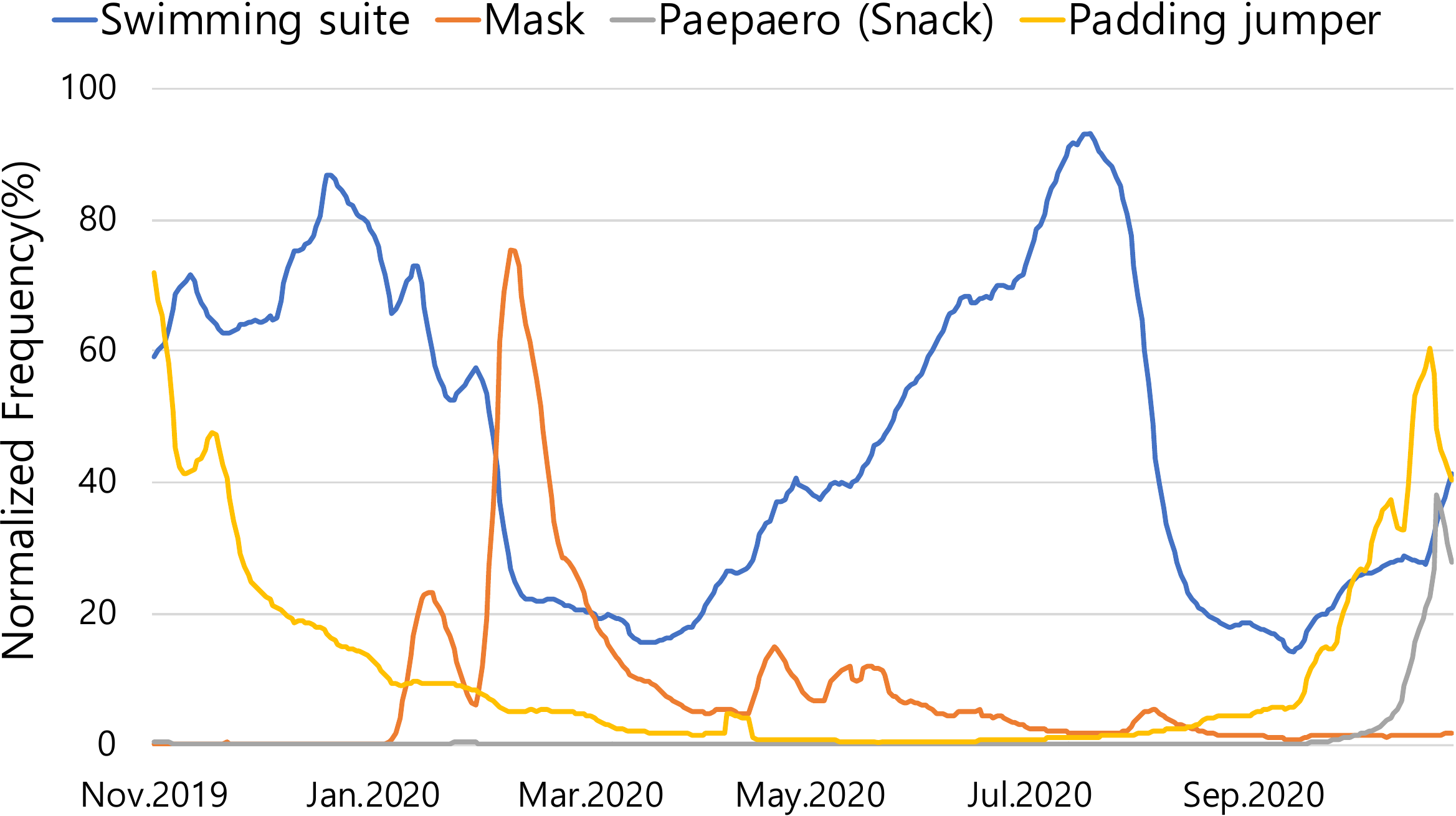}
    \caption{Popularity changes of four items including swimming suite, snack, mask, and padding jumper during one year in a real-world e-commerce service. Each item has its own popularity period and this phenomena is more similar to blurry-CIL than disjoint-CIL because most item categories do not disjointly appear in real-world applications.}
    \vspace{-1em}
    \label{fig:shopping}
\end{figure}

However, they are mostly evaluated in a rather artificial task setup of \emph{disjoint}, where tasks do not share the classes~\cite{parisiKPK18}.
For real-world applications, we consider a more realistic and practical setting of \emph{blurry-CIL} where the classes shared across the tasks~\cite{Prabhu2020GDumbAS} (illustrated at the top of \figurename~\ref{fig:blurry-CIL}).
The blurry-CIL setup requires that (1) each task is given sequentially as a stream, (2) the majority (assigned) classes of tasks differ from each other, and (3) a model can leverage only a very small portion of data of previous tasks.
For instance, suppose an e-commerce service that categorizes new items with their images taken by a seller.
For each category, the number of newly registered items conspicuously depend on various factors such as season and transient event but not reduce to zero. 
The popularity period of items varies according to their characteristics as shown in \figurename~\ref{fig:shopping}; \eg, swimming suits are prevalent in summer and padding jumper are much more registered in winter. 

In recent literature, the methods storing a small portion of old data have shown {promising results} in preserving the information of {old} classes when training new classes for the blurry-CIL setup~\cite{Prabhu2020GDumbAS}, thus alleviating catastrophic forgetting~\cite{goodfellow2013empirical}. This strategy naturally raises the question: \textit{what is the optimal strategies to manage the memory?} Since the number of stored samples is much smaller than that of the incoming new-class, the samples in the memory would incur either overfitting or be ignored during training due to its small size compared to that of samples of incoming tasks. 
As a straightforward solution, if we gradually increase the memory size when the samples are incoming, the problem-setting fails to hold an important resource constraint of the CIL; a limited fixed memory requirement.
Therefore, we need a strategy to maintain sufficient information of the old class with a small number of samples.

To {address} this problem, we investigate two factors for better continual learning on the newly defined \textit{blurry-CIL} setup; \textit{sampling} for the memory and \textit{augmenting} the data in the memory. 
First, we propose a \textit{perturbation-induced uncertainty} to select samples for the memory by measuring the per-sample robustness against the perturbations. 
To measure the uncertainty, we define a prior distribution \jw{that draws} the perturbed samples and approximates the \emph{robustness} (\ie, inverse of uncertainty) described as a likelihood function in a Bayesian formulation.
\jw{We fill the memory slots \jw{with} the samples drawn from the distribution corresponding to the robustness.
We show that the diversity-induced memory by sampling both perturbation-robust and fragile data helps the models to preserve discriminative boundary for each class.}

Second, we investigate the effect of the diversity acquired by data augmentation \jw{in} the blurry-CIL. 
In particular, \jw{label mixing-based data augmentation, such as CutMix~\cite{yun2019cutmix}, projects the input samples into a more complex dimension by mixing the image-label of multiple data samples randomly and has reported notable successes in various recognition tasks~\cite{wu2020generalization,kim2020puzzle}.
It provides additionally rich diversity of stored samples \jw{in the episodic memory.} 
Along with the label mixing augmentation, we exploit the effects of composition of multiple data augmentations for enhancing the diversity, benefiting from conventional methods such as flipping, shearing, or color jittering and recent automated data augmentation researches~\cite{AutoAugment, cubuk2019randaugment,fastauto}.}
Incorporating the two proposals, we name our method as \emph{\fullour} or \emph{\memname} for short.

Our \memname is \yj{mainly} evaluated in blurry-CIL setup on MNIST, CIFAR10, CIFAR100 and ImageNet datasets, compared with various standard CIL methods. 
The extensive experimental validations show that our approach effectively addresses blurry-CIL, outperforming state-of-the art baselines with significant margins. 
In addition, our method comparably performs to the other methods in disjoint-CIL set-up even if it is designed for blurry-CIL setup. 

We summarize the contributions as follows:
\vspace{-0.5em}
\begin{itemize}
\setlength\itemsep{-0.5em}
    \item We propose a new diversity-aware sampling method for effectively managing the memory with limited capacity by leveraging classification uncertainty. 
    \item We propose to augment the samples in the memory to further enhance the diversity of the samples.
    \item Our \memname outperforms previous methods in blurry-CIL setup by large margins.
    \item We release the source code of \memname and the evaluation protocol including the task splits of blurry-CIL for future research in this avenue.
\end{itemize}

\section{Related Work}

\paragraph{Class Incremental Learning Setups.}
Among many scenarios of continual learning, summarized in~\cite{ven2018three}, {we are particularly interested in class-incremental learning (CIL) scenario with no task identity is given at the inference~\cite{gepperthH16}.}
{There have been many proposals  that can be roughly categorized into (1) rehearsal-based approaches~\cite{Chaudhry_2018_ECCV,castro2018eccv,icarl}, where episodic memory stores a few exemplars of old tasks, then the exemplars will be replayed in the future task, and 
(2) regularization-based approaches~\cite{Kirkpatrick2017OvercomingCF,Zenke2017ContinualLT,Liu2018RotateYN,Li2017LearningWF,Lee2017OvercomingCF,Mallya2018PiggybackAA}, where no samples of old tasks is stored, but exploit the information of old tasks implicitly remained in the parameters of models.}
As rehearsal-based approaches generally have shown the better performance in CIL~\cite{Prabhu2020GDumbAS}, we propose to improve memory management and exploit the insufficient information in an episodic memory, presuming the existence of such memory. 

Class-incremental learning usually refers to a sequential learning paradigm with disjoint set of tasks~\cite{icarl,castro2018eccv,gepperthH16}. 
However, recent studies \cite{gss, Prabhu2020GDumbAS} introduce a setup containing blurry and continuous stream of tasks, which is more realistic as many real-world tasks are seldom given in a disjoint manner.
Another setup is whether CIL allows the temporary buffer for storing incoming samples of a current task or not during model training, each of which is called \textit{offline} and \textit{online}, respectively.
Many previous works have been evaluated either of online~\cite{Fini2020OnlineCL, gss, gmed} or offline~\cite{bic, icarl, rwalk, castro2018eccv} setup, while GDumb~\cite{Prabhu2020GDumbAS} reports on both of setups.
Basically, \yj{online is more difficult but more practical,} so we focus on online to report more practical results.
Instead, we investigate the importance of memory management and propose effective memory update algorithm.

\vspace{-1em}\paragraph{Class Imbalance.}
Rehearsal-based approaches have reported severe catastrophic forgetting due to the class-imbalance of exemplars~\cite{bic}. 
This makes models vulnerable to the most frequent classes in episodic memory. 
To address the catastrophic forgetting problem, {GEM~\cite{LopezPaz2017GradientEM}, MER~\cite{riemer2018learning}, and GSS~\cite{gss} propose to update the weights using gradient information so that the models get knowledge from prior task,} 
and BiC~\cite{bic} proposes adding a simple layer at the end of model to calibrate the bias.
Very recently, MEGA~\cite{MEGA} proposes a loss balancing approach mixing loss of previous and current classes to relieve the forgetting. 
HAL~\cite{HAL} proposes a way to utilize the most destructive samples in the past tasks as anchor points to address the forgetting problem, 
and CAL~\cite{CAL} proposes an approach keeping additional information by storing intermediate activations, in addition to the raw images.
However, those approaches overlook the importance of memory management and normally adopt simple random sampling~\cite{MEGA, CAL} or reservoir sampling~\cite{riemer2018learning} or ring-buffer sampling~\cite{HAL}. 


\vspace{-1em}\paragraph{Episodic Memory Management.}
There are a number of strategies proposed in the literature~\cite{parisiKPK18}.
Interestingly, many proposals show marginal accuracy improvement over the uniform random sampling despite the computational complexity~\cite{rwalk,castro2018eccv,icarl}. 
These methods include herding selection~\cite{Welling2009herding}, a discriminative samplings~\cite{mnemonics} and entropy based samples~\cite{Chaudhry_2018_ECCV}. 
The herding selection chooses the samples proportional to a histogram of each sample's distance to the class mean.
The discriminative sampling chooses the samples that define decision boundaries.
The entropy based sampling method chooses the samples by the entropy of their softmax distribution in the output layer.

To obtain the representative and discriminative exemplars, Liu~\etal proposes a complex but effective sampling method guaranteeing that the exemplars well represents the mean and boundary of each class distribution~\cite{mnemonics}.
Also, Borsos~\etal propose a coreset generation method for the representative memory using cardinality-constrained bi-level optimization~\cite{Borsos2020CoresetsVB}.
and Cong \etal propose a GAN based memory which they can perturb styles of remembered samples for incremental learning~\cite{Cong2020GANMW}.
These recently published works address the quality of the samples stored in the memory, they are either computationally expensive or difficult to train a sample generator for the memory~\cite{Borsos2020CoresetsVB}.

Other than sampling, there are works addressing the episodic memory. 
Generative models are employed to generate past task samples~\cite{Shin2017ContinualLW,Seff2017ContinualLI,Wu2018MemoryRG,hu2018overcoming} instead of sampling. 
The generation strategy is an active research topic and shows promising results in relatively straightforward experimental validation (\eg, on MNIST and SVHN).
But on these datasets, sampling from the uniform distribution already achieves saturated accuracy~\cite{Chaudhry_2018_ECCV} and there is no promising results reported in challenging datasets (\eg, ImageNet) yet.
Hayes \etal propose to replay `compressed memory' to increase the memory utilization~\cite{Hayes2020REMINDYN}.
Iscen \etal propose to reduce the dimension of stored features for efficiency~\cite{Iscen2020MemoryEfficientIL}.
Fini \etal propose a batch-level distillation (BLD) method to increase the memory efficiency in an online setting which has an extreme memory constraint~\cite{Fini2020OnlineCL}.
Unlike these works addressing the sampling efficiency, we focus on the quality of the stored samples in the memory.

\section{Class Incremental Learning Setups}
\label{sec:problem_def} 


We can formulate CIL setups as follows:
\begin{equation*}
\begin{split}
    C &= \{c_{1}, c_{2}, \ldots , c_{N}\}, \\
    T_{t} &= \{c~|~\psi(c) = t \}, \\
    \mathfrak{D}^{C}_{c} &= \{x^{c}_{1}, x^{c}_{2}, \ldots, x^{c}_{M_{c}}\}, \\
    \mathfrak{D}^{T}_{t} &= \{\mathfrak{D}^{C}_{c}~|~c \in T_{t}\}, \\
\end{split}
\end{equation*}
where $C$ denotes a set of all classes, $T_{t}$ denotes a class-subset assigned to each task $t$, which is determined by a stochastic assign function, $\psi(c)$ returning an assigned task for a given class $c$, and $\mathfrak{D}^{C}_{c}$ and $\mathfrak{D}^{T}_{t}$ represent a set of samples populating class $c$ and task $t$ sample space, respectively.
Note that $N$ is not known and not even bounded in real-world scenario and $M_{c}$ can be either of equal or not among classes ($c$) according to a problem definition.

We now formulate either blurry or disjoint CIL setups by intersecting $\mathfrak{D}^{T}_{t}$'s or not. 
\begin{equation*}
\begin{split}
    \text{disjoint-CIL} \Rightarrow \bigcap{{T}_{t}} = \varnothing,\\
    \text{blurry-CIL} \Rightarrow \bigcap{{T}_{t}} \neq \varnothing.
\end{split}
\end{equation*}

The disjoint-CIL setup exaggerates the catastrophic forgetting since it never exposes seen classes in successive tasks, but it is deviated from the real-world where new classes do not show up exclusively.
Conversely, blurry-CIL setup makes the task boundaries faint in a way that each task contains small number of classes also present in the other tasks. 
Approaches are evaluated in various perspectives including forgetting and intransigence~\cite{rwalk} under a continuously changing class balance setup~\cite{Prabhu2020GDumbAS}.

\section{Approach}
\label{sec:approach}

To effectively address the blurry-CIL with an episodic memory, we propose a memory management strategy that enhances diversity of samples to cover the distribution of the class by sampling a diverse set of samples which may preserve the boundary of a class distribution.
We further enhance the diversity of the samples by data augmentation. 

\subsection{Diversity-Aware Memory Update}\label{sec:diversity_aware_memory_update}
We argue that the exemplars which are selected to be stored in the memory should be not only representative for their corresponding class but also discriminative to the other classes.
To choose such samples, we argue that the samples that are near the classification boundary are the most discriminative and the samples that are close to the center of the distribution is the most representative.
To satisfy both characteristics, we propose to sample the exemplars that are \emph{diverse} in the feature space.

To secure the diversity, we need to estimate the relative locations of each sample in class-discriminative feature space.
But it is computationally expensive to compute the relative locations of the features as it requires to compute sample-to-sample distances ($O(N^2)$).
Instead, we propose to estimate the relative location by \emph{uncertainty} of a sample estimated by the classification model, \ie, we assume that the more certain samples for the model will be located closer to the center of the class distribution and \emph{vice versa}.

\begin{figure}[t!]
    \centering
    \includegraphics[width=1\columnwidth]{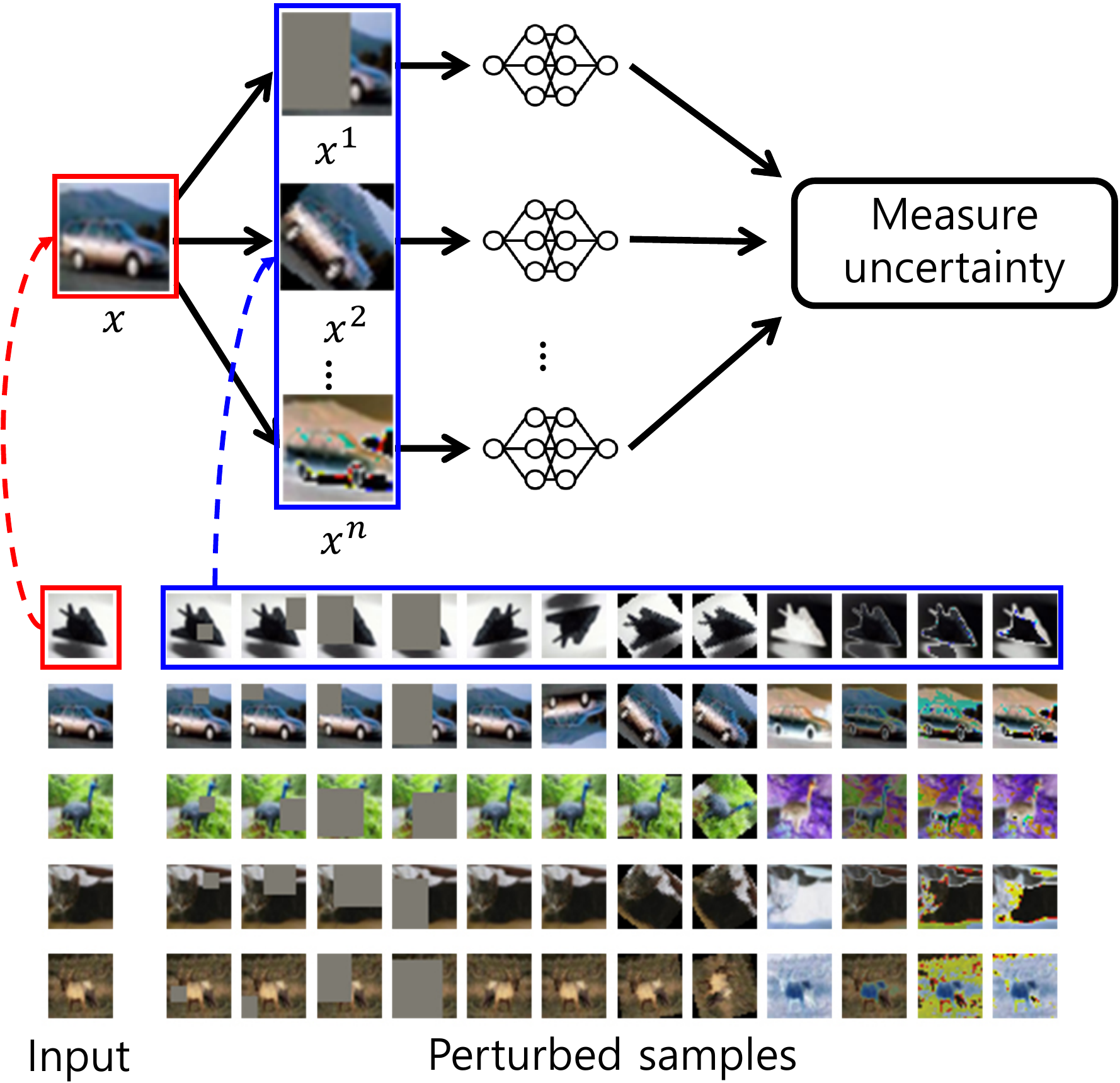}
    \vspace{-3mm}
    \caption{Estimating uncertainty of a data sample ($x$) with its perturbed samples ($\tilde{x}$) for the proposed \fullour. Detailed procedure is summarized in Algorithm~\ref{alg:memory_update}.}
    \vspace{-1em}
    \label{fig:uncertainty}
\end{figure}

Specifically, we compute uncertainty of a sample by measuring the variance of model outputs of perturbed samples by various transformation methods for data augmentation: including color jitter, shear, and cutout~\cite{devries2017cutout} (illustrated in \figurename~\ref{fig:uncertainty}).
Following the derivation from Gal~\etal~\cite{gal2016dropout}, we approximate the uncertainty by Monte-Carlo (MC) method of the distribution $p(y=c|x)$ when given the prior of the perturbed sample $\tilde{x}$, as $p(\tilde{x}|x)$. 
We define the perturbation prior $p(\tilde{x}|x)$, as a uniform mixture of the various perturbations as shown in the examples in \figurename~\ref{fig:uncertainty}. 
The derivation can be written as:
\begin{equation}
\label{eq:monte_carlo}
\begin{split}
    p(y=c|x) &= \int_{\mathcal{\tilde{D}}}{p(y=c|\tilde{x}_{t}) p(\tilde{x}_{t}|x) d\tilde{x}_{t}} \\
    & \approx \frac{1}{A} \sum_{t=1}^{A}{p(y=c|\tilde{x}_{t})},
\end{split}
\end{equation}
where $x$, $\tilde{x}$, $y$ and $A$ denote a sample, a perturbed sample, the label of the sample, and the number of perturbation methods, respectively.
The distribution $\mathcal{\tilde{D}}$ denotes the data distribution defined by the perturbed samples $\tilde{x}$.
In particular, the perturbed sample $\tilde{x}$ is drawn by a random function $f_r(\cdot)$, as:
\begin{equation}
\begin{split}
    \label{eq:perturb}
    \tilde{x} = f_{r}(x | \theta_r), r=1,...,R,
\end{split}
\end{equation}
where $\theta_r$ is a hyper-parameter which denotes the random factor of the $r$-th perturbation. 
The prior $p(\tilde{x}|x)$ is defined as:
\begin{equation}
\begin{split}
    \label{eq:purturb1}
    \tilde{x} \sim \sum^{R}_{r=1} w_r*f_{r}(x | \theta_r),
\end{split}
\end{equation}
where the random variable $w_r, r=\{1,...,R\}$ is drawn from a categorical binary distribution.
From the approximated distribution \eqref{eq:monte_carlo}, we measure the uncertainty of the sample with respect to the perturbation as:
\begin{equation}
\begin{split}
    \label{eq:var_ratio}
    S_{c} &= \sum_{t=1}^{T} \mathbbm{1}_{c}{\argmax_{\hat{c}}{p(y=\hat{c}|\tilde{x}_{t})}},\\
    u(x) &= 1 - \frac{1}{T} \max_{c}{S_{c}},
\end{split}
\end{equation}
where $u(x)$ denotes the uncertainty of the sample $x$ and $S_c$ is the number of times that class $c$ is the predicted top-1 class. 
The $\mathbbm{1}_{c}$ denotes the binary class indexing vector.
The lower valued $u(x)$, corresponding to more consistent top-1 class over perturbations, indicates that $x$ resides in a region where a model is strongly confident.

Algorithm~\ref{alg:memory_update} summarizes our proposed diversity-aware memory update algorithm. 
Following GDumb~\cite{Prabhu2020GDumbAS}, we also assign the same amount of memory slots ($k_{c}$) over the `seen' classes ($N$).
After assigning the exemplars to the memory slots, we compute the uncertainties for both streamed samples ($\mathfrak{D}^{S}_{t}$) and stored exemplars ($\mathfrak{D}^{M}_{t-1}$) in a memory at task $t$, then sort all these samples ($\mathfrak{D}_c$) by their uncertainties.
From the sorted list, we select samples with an interval of $|\mathfrak{D}_c| / k_c$  to secure the diversity.
As a result of this sampling, we fill the memory with exemplars in a wide spectrum ranged from strongly perturbation, \ie, robust samples, to fragile ones.
This imposes perturbation-based diversity to the episodic memory.

\definecolor{azure}{rgb}{0.0, 0.5, 1.0}
\begin{algorithm}[t!]
    \caption{Diversity-Aware Memory Update} 
	\label{alg:memory_update}
	\begin{algorithmic}[1]
	    \State {\textbf{Input:} $K$ denotes memory size, $N_{t}$ denotes the number of seen classes until task $t$, $\mathfrak{D}^{S}_{t}$ denotes stream data at task $t$, $\mathfrak{D}^{M}_{t-1}$ denotes exemplars stored in a episodic memory after task $t-1$.} 
	    \State {\textbf{Output:} $\mathfrak{D}^{M}_{t}$ exemplars after learning task $t$.}
		\State {$\mathfrak{D}^{M}_{t}$ = \{\} \quad {\color{azure} \Comment{\small New exemplars from scratch}}}
		\State {$k_c = floor(K / N_t) $ \quad {\color{azure} \Comment{\small Class-balanced sampling}}} 
		\For {$c=1,2,\ldots,N_t$}
		    \State {$\mathfrak{D}_c = \{(x,y) | y=c, (x,y) \in \mathfrak{D}^{S}_{t} \cup \mathfrak{D}^{M}_{t-1}$\} }
		    \State {Sort $\mathfrak{D}_c$ by $u(x)$ computed by~\eqref{eq:var_ratio}}
		    \For {$j = 1,2, \ldots, k_c$}
		    \State {$i = j * |\mathfrak{D}_c| / k_c$ \quad {\color{azure} \Comment{\small $|\mathfrak{D}_c| / k_c$ step-size indexing}}}
		    \State {$\mathfrak{D}^{M}_{t}$  += $\mathfrak{D}_c[i]$}
		    
		\EndFor
	\EndFor
	
	\end{algorithmic} 
\end{algorithm}

\subsection{Diversity Enhancement by Augmentation}\label{sec:data_augmentation}
To further enhance the diversity of exemplars from the memory, we employ data augmentation (DA).
The DA's diversify a given set of samples by image-level or feature-level perturbations, which correspond to the philosophy of updating memory by securing the diversity (Section~\ref{sec:diversity_aware_memory_update}).

We consider various perturbation types including simple single-image-based DA perturbing the original input image, mixed-labeled DA which integrates multiple images~\cite{yun2019cutmix,zhang2017mixup} and automated DAs (AutoDAs)~\cite{AutoAugment, cubuk2020randaugment,fastauto}.
The stochastically chosen various augmentations succeed in image classification.
Yet, the efficacy of the DA's has not been well investigated in the CIL context.

\vspace{-1em}\paragraph{Mixed-Label Data Augmentation.}
As task iteration proceeds, the samples in a new task are likely to follow different distribution from the one in the episodic memory (\ie, from the previous tasks). 
We adopt mixed-labeled DA to `mix' images in the classes of the new tasks and the exemplars of the old classes in the memory. 
This mixed-label DA alleviates the side effects caused by change of class distribution over the tasks and improves the performances. 

As one of the representative mixed-labeled DA methods, CutMix~\cite{yun2019cutmix} generates a mixed sample and a smoothed label, given the set of supervised samples $(x_1, y_1)$ and $(x_2, y_2)$, as:
\begin{equation}
\begin{split}
    \Tilde{x} & =  \mathbf{m} \odot x_{1} + (\mathbf{1}- \mathbf{m}) \odot x_{2}, \\
    \Tilde{y} & =  \lambda y_1 + (1-\lambda) y_2, \\
\end{split}
\label{eq:cutmix}
\end{equation}
where the set $\mathbf{m}$ denotes the randomly selected pixel region for the image $x_1$ according to the hyper-parameter $\beta$ drawn from the beta-distribution.
As shown in \eqref{eq:cutmix}, the mixed-label DA generates artificial samples that are hard to be considered as a variant of the source images unlike the conventional data augmentations manipulating an original image by flipping, rotating, and/or contrasting while not ruining a class boundary.

\vspace{-1em}\paragraph{Automated Data Augmentation.}
In addition to the above mixed-labeled DAs, we further use AutoDA to enrich the augmentation effect by compositing multiple DA's on the model performance under CIL.
Especially, 
\jw{we employ AutoAugment~\cite{AutoAugment}, providing parameters for determining the number of augmentations and their magnitudes.}




\section{Experiments}\label{sec:experiments}
\jw{We empirically validate the efficacy of our \memname by comparing it with state of the arts in various experimental setups; CIL task setups for the benchmarks, memory-sizes of episodic memory, and performance metrics.
In addition, we further investigate components of the propose \memname; memory management strategy and augmentation methods for their contribution to the CIL performances.}

\subsection{Experimental Setup}
\paragraph{Benchmark Task Setup.}
We evaluate algorithms mostly in blurry-CIL setup, otherwise stated.
Following~\cite{gss}, we denote blurry-CIL setup as `Blurry$M$', where the $M$ denotes the portion of samples coming from the other tasks. 
Therefore, each task in the blurry-CIL setup contains samples from its assigned major classes (\ie, the most frequent classes and assigned to each task exclusively)  consisting of $(100-M)\%$ and ones of minor classes (\ie, the other classes of $C$ except for the assigned major classes) consisting of remaining $M\%$.
Note that the class distribution of minor classes in each task are balanced.

In addition, we consider two different learning setup; \textit{online} and \textit{offline}. 
In online, the incoming samples are presented to a model only once except the ones selected as exemplars since it does not have a buffer which is large enough to keep the whole streamed samples.
\jw{On the other hand, in offline, a model can observe the incoming samples multiple times (\ie, epochs) with the buffer.}  
Please note that we repeat each experiment three times to report means and standard deviations except the ImageNet experiments. 

\vspace{-1em}\paragraph{Datasets and Metrics.}
We use MNIST, CIFAR10, CIFAR100 and ImageNet (ILSVRC2012) datasets to configure CIL task setups for evaluations. 
We randomly split and assign with different random seeds a set of all classes ($C$) into 5 tasks to generate a CIL task setup, and thus each class-subset ($T_t$) has 2 and 20 major classes for CIFAR10 and CIFAR100 datasets, respectively. 
For ImageNet, we split 1000 classes to 10 tasks, so each class-subset ($T_t$) has 100 major classes.

We use three popular metrics in the literature, such as \textit{Last Accuracy (A5)}, \textit{Last Forgetting (F5)}, and \textit{Intransigence (I5)}.
`Last' refers to the value is measured after all tasks are learned, and we denote it with number `5' here because both of CIFAR10 and CIFAR100 have five tasks. 
Accordingly, they will be A10, F10, and I10 for ImageNet.
Please refer to the supplementary material for more details about the metrics.
Finally, we use various episodic memory sizes for different datasets as the size of the datasets differ.

\vspace{-1em}\paragraph{Baselines and Implementation Details.} We compare our proposed \memname with the standard CIL methods including EWC~\cite{ewc}, Rwalk~\cite{rwalk}, iCaRL~\cite{icarl}, BiC~\cite{bic} and GDumb~\cite{Prabhu2020GDumbAS}, the only method specifically designed for the blurry setup. 
Note that GSS~\cite{gss} is not compared since GDumb outperforms it by large margins.
The comparable CIL methods utilize MLP400, ResNet18, ResNet32, and ResNet34~\cite{resnet} as their network architectures for MNIST, CIFAR10, CIFAR100, and ImageNet, respectively.
For CIFAR10/100, we use the same backbone to the official GDumb~\cite{Prabhu2020GDumbAS} implementation\footnote{https://github.com/drimpossible/GDumb} \jw{throughout all experiments.}
For ImageNet, we use the backbone from their original implementation~\cite{resnet}.

For the training hyperparameters of experiments on MNIST and CIFAR10/100, we use batch-size of 16, cosine annealing learning-rate schedule ranged from 0.05 to 0.0005, and the number of epochs of 256, following~\cite{Prabhu2020GDumbAS}.  
For those on ImageNet, we use batch-size of 256, step annealing learning-rate schedule ranged from 0.1 to 0.001, and the number of epochs of 100, which are used from BiC~\cite{bic}.

In addition, we use an episodic memory, which is updated through reservoir sampling which exhibits the best performance (Section~\ref{subsec:RMU}), to the baselines not considering the existence of memory; EWC and Rwalk, for fair comparison.
As expected, all memory-attached baselines outperform the corresponding original ones.

\subsection{Results} 

\begin{table*}[]
\centering
\caption{Comparison with three metrics (A\{5, 10\}, F\{5, 10\}, and I\{5, 10\}: \%) in \{MNIST, CIFAR100, ImageNet\}-Blurry10-Online. 
\newline $^*$ indicates the reproduction of BiC with only using classification loss without distilling loss to be better suited for Blurry10 setup.}
\label{tab:various-dataset-blurry10-online}
\resizebox{\linewidth}{!}{%
\begin{tabular}{@{}crrrrrrlll@{}}
\toprule 
& \multicolumn{3}{c}{MNIST (K=500)} & \multicolumn{3}{c}{CIFAR100 (K=2,000)}  & \multicolumn{3}{c}{ImageNet (K=20,000)} \\ 
Methods & \multicolumn{1}{c}{A5 ($\uparrow$)} & \multicolumn{1}{c}{F5 ($\downarrow$)} & \multicolumn{1}{c}{I5 ($\downarrow$)} & \multicolumn{1}{c}{A5 ($\uparrow$)} & \multicolumn{1}{c}{F5 ($\downarrow$)} & \multicolumn{1}{c}{I5 ($\downarrow$)} & \multicolumn{1}{c}{A10 ($\uparrow$)} & \multicolumn{1}{c}{F10 ($\downarrow$)} & \multicolumn{1}{c}{I10 ($\downarrow$)} \\ 
\cmidrule(lr){1-1} \cmidrule(lr){2-4} \cmidrule(lr){5-7} \cmidrule(lr){8-10} 

EWC & 90.98 $\pm$ 0.61 & 4.23 $\pm$ 0.45 & 4.54 $\pm$ 0.94 & 26.95 $\pm$ 0.36 & 11.47 $\pm$ 1.26 & 43.18 $\pm$ 14.22 & 39.54 & 14.41 & 42.68 \\
Rwalk & 90.69 $\pm$ 0.62 & 4.77 $\pm$ 0.36 & 4.96 $\pm$ 0.56 & 32.31 $\pm$ 0.78 & 15.57 $\pm$ 2.17 & 37.18 $\pm$ 10.02 & 35.26 & 13.92 & 46.96 \\
iCaRL & 78.09 $\pm$ 0.60 & 6.09 $\pm$ 0.23 & 17.03 $\pm$ 0.60 & 17.39 $\pm$ 1.04 & 5.38 $\pm$ 0.88 & 44.18 $\pm$  9.29 & 17.52 & 1.94 & 81.94 \\
GDumb & 88.51 $\pm$ 0.52 & 2.67 $\pm$ 0.31 & 6.75 $\pm$ 0.43  & 27.19 $\pm$ 0.65 & 7.49 $\pm$ 0.95 & 41.18 $\pm$  7.23 & 21.52 & 4.07 & 60.70 \\
BiC & 77.75 $\pm$ 1.27 & 8.25 $\pm$ 1.45 & 17.37 $\pm$ 1.27 & 13.01 $\pm$ 0.24 & 4.63 $\pm$ 0.46 & 53.84 $\pm$ 11.85 & 37.20$^*$ & 1.52$^*$ & 45.02$^*$ \\
\cmidrule(lr){1-10} 
\textbf{\memname w/o DA} & \textbf{92.65 $\pm$ 0.33} & \textbf{0.58 $\pm$ 0.09} & \textbf{3.14 $\pm$ 0.94} & 34.09 $\pm$  1.41 & \textbf{4.01 $\pm$ 0.50} & 34.51 $\pm$ 4.58 & 37.96 & 2.63 & 44.26 \\
\textbf{\our} & 91.80 $\pm$ 0.69 & 0.75 $\pm$ 0.30 & 3.62 $\pm$ 0.63 & \textbf{41.35 $\pm$ 0.95} & 4.99 $\pm$  0.89 & \textbf{20.18 $\pm$ 3.06} & \textbf{50.11} & \textbf{1.39} & \textbf{32.11} \\ \bottomrule
\end{tabular}%
}

\end{table*}

\begin{table*}[t]
\centering
\caption{Comparison with three metrics (A5, F5, and I5: \%) for three episodic memory sizes in CIFAR10-Blurry10-Online. DA is used in \our denotes CutMix+AutoAug.}
\label{tab:cifar10-blurry10-online}
\resizebox{\linewidth}{!}{%
\begin{tabular}{@{}crrrrrrrrr@{}}
\toprule
& \multicolumn{3}{c}{ { K=200}} & \multicolumn{3}{c}{ { K=500}} & \multicolumn{3}{c}{ { K=1,000}} \\
Methods & A5 ($\uparrow$) & F5 ($\downarrow$) & I5 ($\downarrow$) & A5 ($\uparrow$) & F5 ($\downarrow$) & I5 ($\downarrow$) & A5 ($\uparrow$) & F5 ($\downarrow$) & I5 ($\downarrow$) \\ 
\cmidrule(lr){1-1} \cmidrule(lr){2-4} \cmidrule(lr){5-7} \cmidrule(lr){8-10}
EWC & 40.07 $\pm$ 2.14 & 21.20 $\pm$ 0.76  & 61.91 $\pm$ 4.51 & 55.65 $\pm$ 4.60 & 16.06 $\pm$ 3.89 & 44.24 $\pm$ 11.98 & 68.67 $\pm$ 0.95 & 12.63 $\pm$ 1.78 & 25.97 $\pm$ 10.88 \\
Rwalk & 38.66 $\pm$ 1.52 & 20.67 $\pm$ 2.36 & 65.81 $\pm$ 4.85 & 53.66 $\pm$ 3.18 & 17.04 $\pm$ 0.31 & 45.81 $\pm$ 9.78 & 68.20 $\pm$ 1.86 & 11.48 $\pm$ 1.19 & 25.17 $\pm$ 11.57 \\
iCaRL & 37.43 $\pm$ 1.31 & 2.08 $\pm$ 2.23 & 63.51 $\pm$ 13.73 &  45.98 $\pm$ 3.04 & 4.75 $\pm$ 1.70 & 51.91 $\pm$ 2.57 & 53.60 $\pm$ 2.82 & 7.21 $\pm$ 2.58 & 37.84 $\pm$ 13.49 \\
GDumb & 35.85 $\pm$ 1.03 & 1.67 $\pm$ 3.49 &  55.31 $\pm$ 6.02 & 49.47 $\pm$ 1.08 & 1.44 $\pm$ 2.77 & 40.91 $\pm$ 14.04 & 64.26 $\pm$ 1.21 & 0.37 $\pm$ 1.92 & 31.81 $\pm$ 13.37\\
BiC & 33.29 $\pm$ 0.86 & 3.91 $\pm$ 1.64 & 50.37 $\pm$ 6.96  & 42.06 $\pm$ 2.41 & 1.34 $\pm$ 2.27 & 52.04 $\pm$ 15.50 & 47.81 $\pm$ 3.04 & 3.03 $\pm$ 1.44 & 52.77 $\pm$ 15.54 \\ 
\cmidrule(lr){1-10} 
\textbf{\memname w/o DA} & 44.41 $\pm$ 1.40 & 0.90 $\pm$ 0.93 & 49.51 $\pm$ 11.09  & 60.87 $\pm$ 0.88 & 0.95 $\pm$ 1.14 & 35.74 $\pm$ 13.89 & 70.93 $\pm$ 1.57 & \textbf{-1.43 $\pm$ 0.71} & 22.07 $\pm$ 14.07 \\
\textbf{\memname} & \textbf{54.61 $\pm$ 1.62} & \textbf{-2.60 $\pm$ 1.91} & \textbf{43.57 $\pm$ 11.63} & \textbf{71.13 $\pm$ 0.25} & \textbf{-0.85 $\pm$ 0.28} & \textbf{18.29 $\pm$ 14.21} & \textbf{78.04 $\pm$ 0.50} & 1.29 $\pm$ 1.26 & \textbf{11.64 $\pm$ 5.83}\\ \bottomrule
\end{tabular}
}
\end{table*}

We compare the propose \memname to other methods in `Blurry10-Online' setup on various datasets and summarize the results in \tablename~\ref{tab:various-dataset-blurry10-online}. 
As shown in the table, \memname consistently outperforms all other methods, and the gain becomes larger when the number of classes ($|C|$) increases, which is more challenging.
Note that the original BiC performs significantly worse in ImageNet in the blurry setup, so we eliminate the distilling loss yielding irregular values, then BiC performs reasonably well (denoted by $^*$ in \tablename~\ref{tab:various-dataset-blurry10-online}). 
On MNIST, however, \memname without DA performs the best. 
We believe that DA interferes the model training with perturbed samples since the exemplars are enough to avoid forgetting.
On the other hand, DA improves the metrics with large margins on the other datasets as we expected in section~\ref{sec:data_augmentation}.

\tablename~\ref{tab:cifar10-blurry10-online} presents the comparison on CIFAR10-Blurry10-Online for three episodic memory sizes ($K$); 200, 500 and 1,000.
We again observe that our proposed \memname outperforms all other baselines over all three memory-sizes in terms of A5, F5, and I5 by significant margins in Blurry and online CIL setup on CIFAR10. 
It is interesting that EWC and Rwalk do not perform well in forgetting (F5) despite their competitive A5 scores regardless of the memory size. 
The results imply that these methods preserve effective exemplars in the final task, which are enough to restore the forgetting happening in the previous tasks.
iCaRL, GDumb and BiC are less effective for intransigence (I5) with larger memory size while they perform well in forgetting compared to EWC and Rwalk as a tradeoff. 

Our \memname not only outperforms other baselines for accuracy but also exhibit good forgetting and intransigence performance, regardless of memory sizes.
It is also observed that the performance gaps between ours and the others decrease when the memory-size becomes larger since the impact of sampling efficiency decreases with redundant samples.
%
Note that these results on CIFAR10 exhibit similar trends to the results on CIFAR100 and ImageNet (shown in \tablename~\ref{tab:various-dataset-blurry10-online}). 
Although the CIFAR100 and ImageNet has 10$\times$ or 100$\times$ more classes than the CIFAR10, \memname still outperforms all the baselines in all three metrics by large margins. 
These results imply that our \memname is quite effective for more practical and realistic CIL setup of blurry and online, compared to the prior arts.

\begin{figure*}[t!]
    \centering
    \begin{subfigure}[t]{0.245\textwidth}
        \includegraphics[width=0.99\columnwidth]{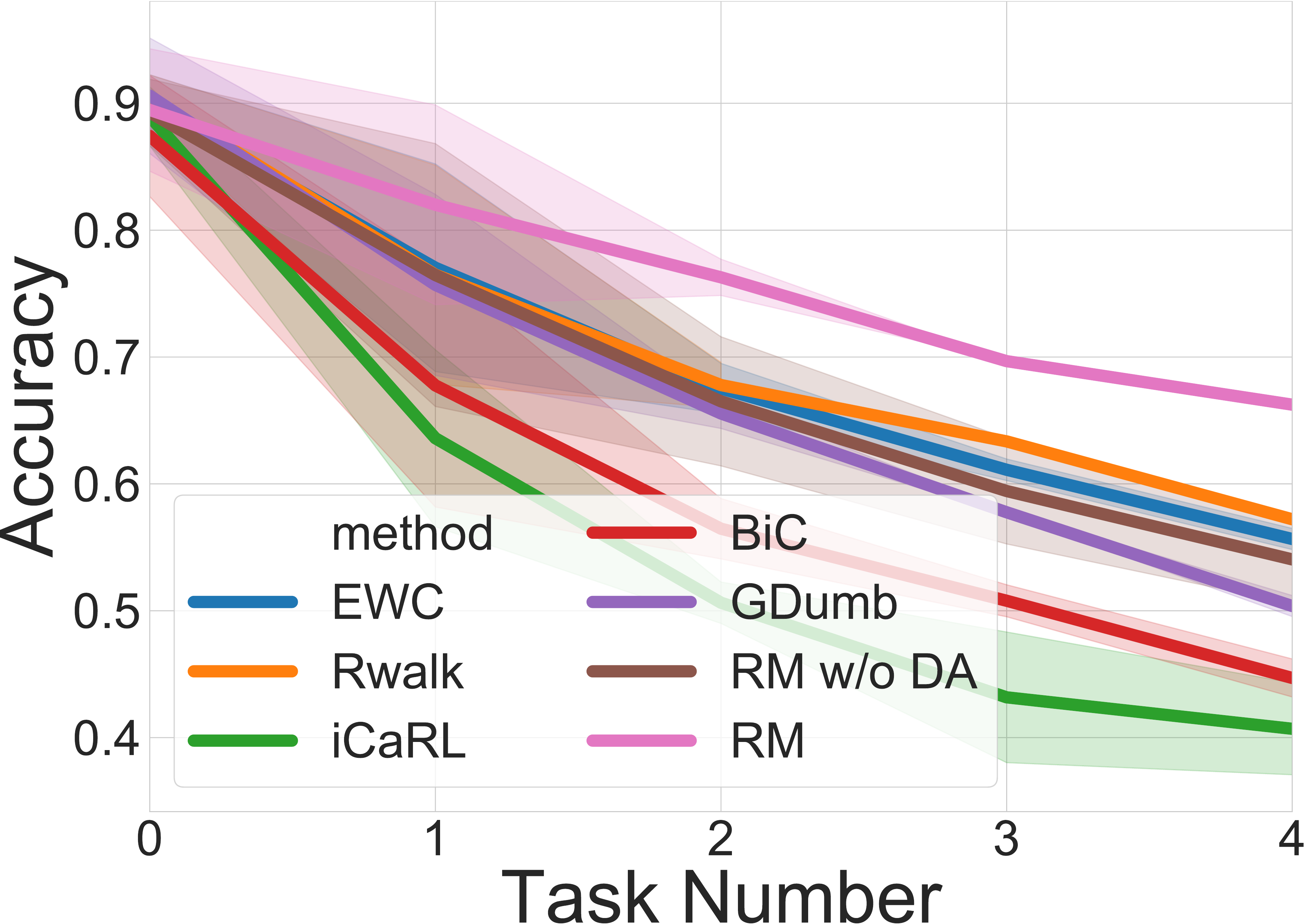}
        \caption{CIFAR10-Disjoint-Online}
        \label{fig:online_disjoint}
    \end{subfigure}
    \begin{subfigure}[t]{0.245\textwidth}
        \includegraphics[width=0.99\columnwidth]{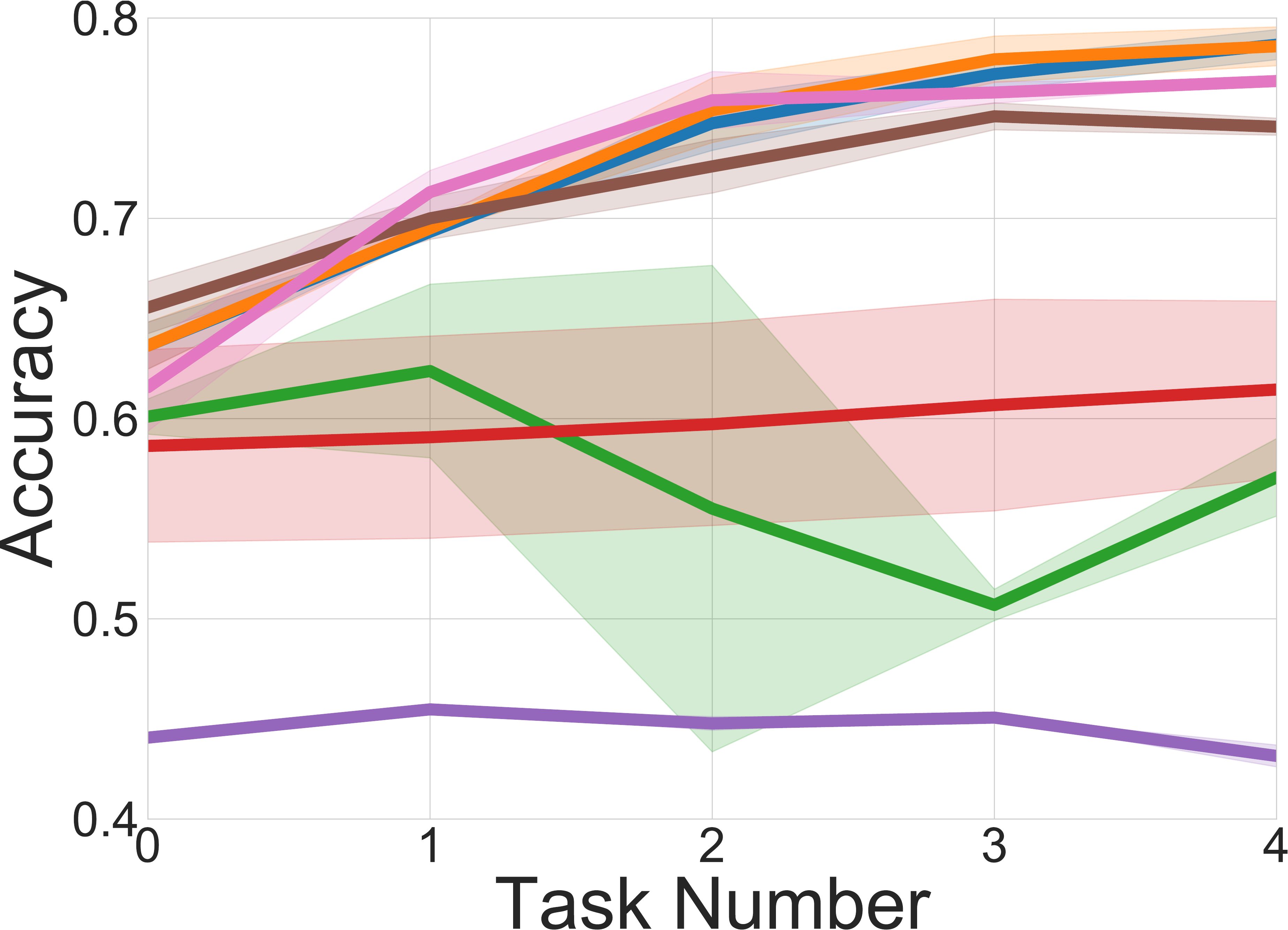}
        \caption{CIFAR10-Blurry10-Offline}
        \label{fig:offline_blurry}
    \end{subfigure}
    \begin{subfigure}[t]{0.245\textwidth}
        \includegraphics[width=0.99\columnwidth]{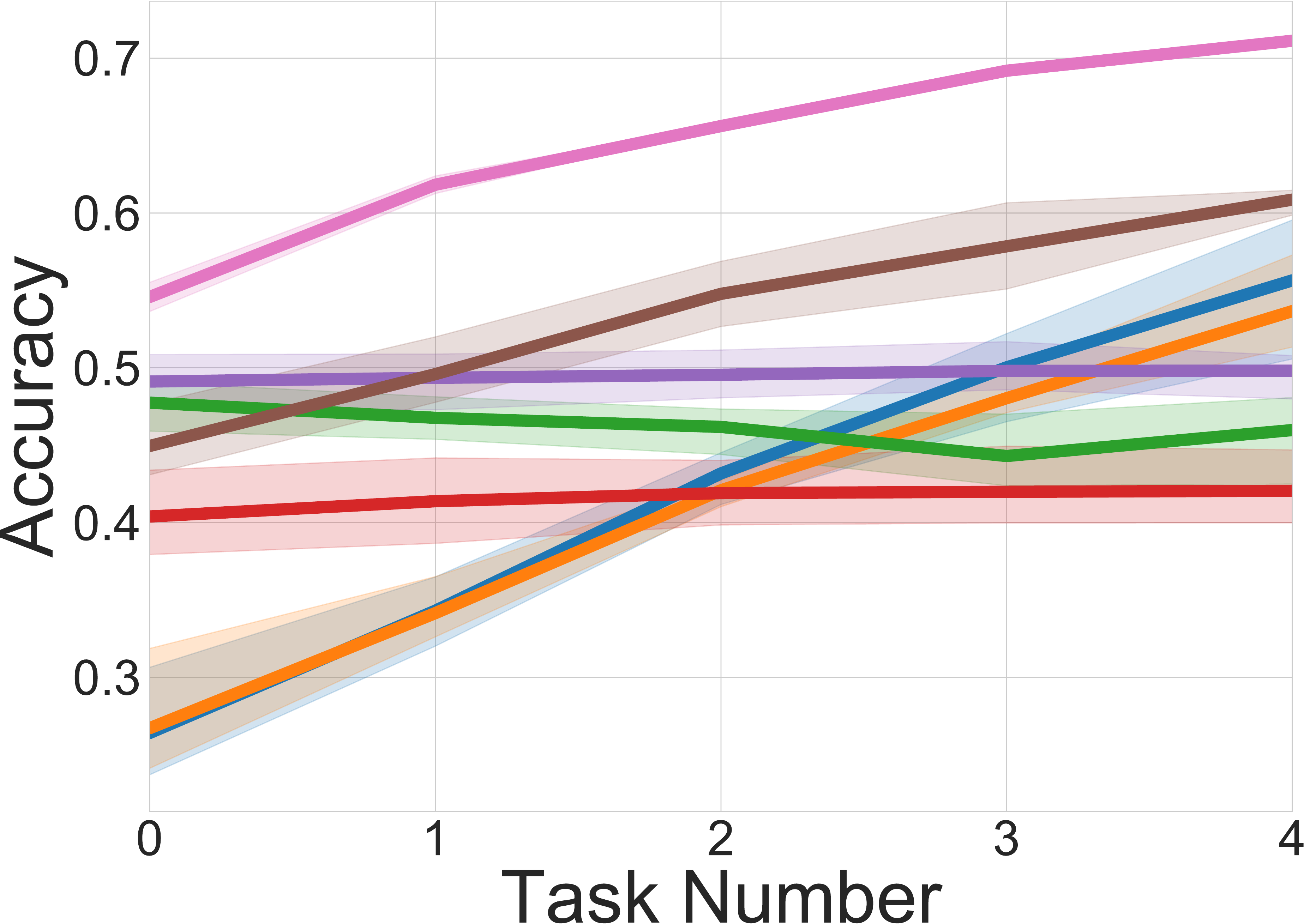}
        \caption{CIFAR10-Blurry10-Online}
        \label{fig:online_blurry}
    \end{subfigure}
    \begin{subfigure}[t]{0.245\textwidth}
        \includegraphics[width=0.99\columnwidth]{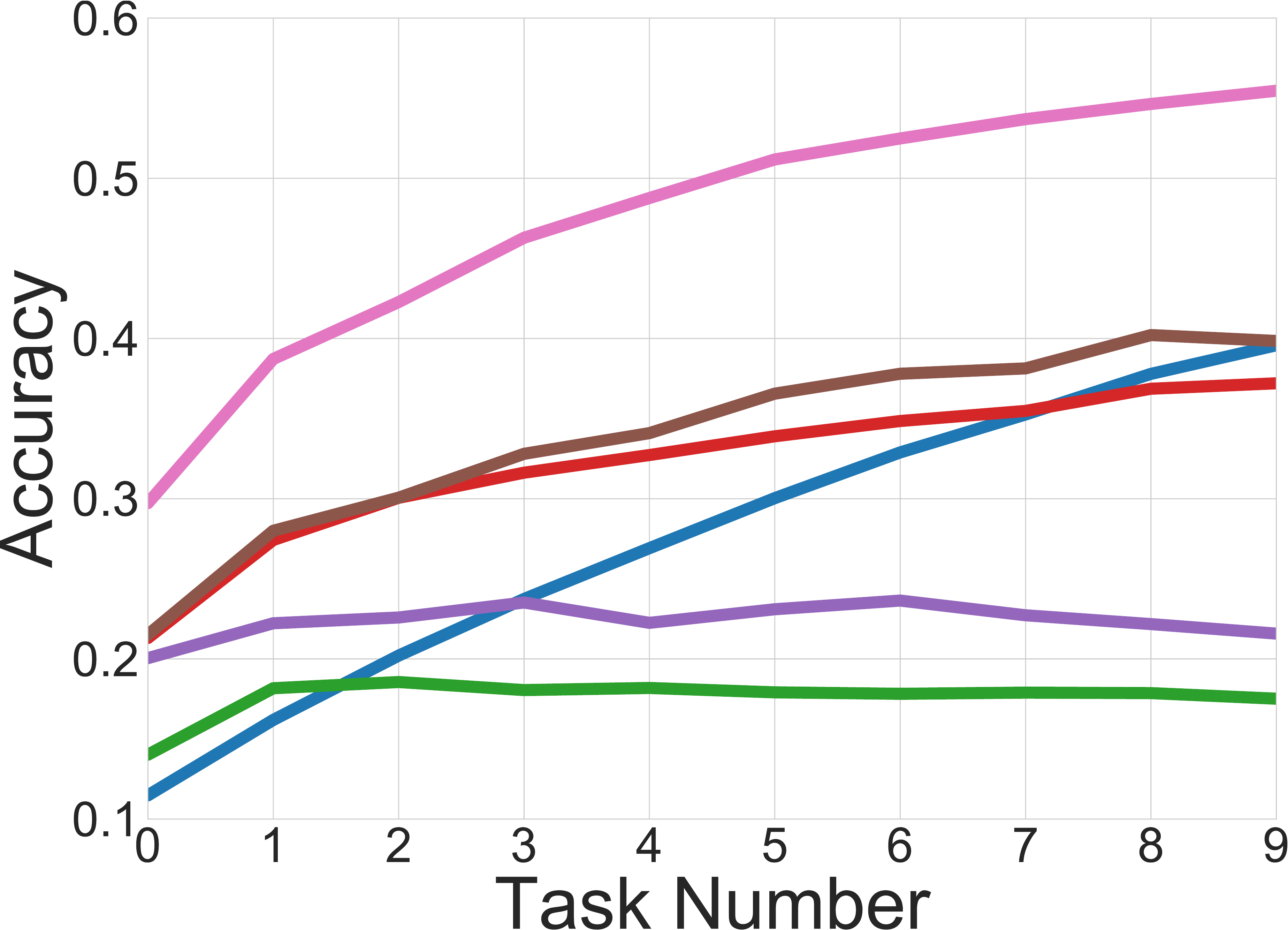}
        \caption{ImageNet-Blurry10-Online}
        \label{fig:imgnet_online_blurry}
    \end{subfigure}
    

    \caption{Illustration of accuracy changes as tasks are being learned in
    (a) CIFAR10-Disjoint-Online, 
    (b) CIFAR10-Blurry10-Offline, 
    (c) CIFAR10-Blurry10-Online,
    (d) ImageNet-Blurry10-Online settings. More results are presented in the supplement.}
    \label{fig:allgraphs}
\vskip -0.1in
\end{figure*}

\subsection{Detailed Analysis}

\begin{table*}[]
\centering
\caption{Comparison of last accuracy (A5 ($\uparrow$), \%) over benchmarks \{Disjoint (0\%), Blurry (10\%), and Blurry (30\%)\} and training setups \{Online and Offline\} on CIFAR10 (K=500).}
\label{tab:cifar10-blurry-disjoint}
\resizebox{0.8\linewidth}{!}{%
\begin{tabular}{@{}crrrrrr@{}}
\toprule
 & \multicolumn{2}{c}{Blurry0 (=Disjoint)} & \multicolumn{2}{c}{Blurry10} &  \multicolumn{2}{c}{Blurry30} \\
Methods & \multicolumn{1}{c}{Online} & \multicolumn{1}{c}{Offline} & \multicolumn{1}{c}{Online} & \multicolumn{1}{c}{Offline}  & \multicolumn{1}{c}{Online} & \multicolumn{1}{c}{Offline} \\ 
\cmidrule(lr){1-1} \cmidrule(lr){2-3} \cmidrule(lr){4-5} \cmidrule(lr){6-7}
EWC               & 55.66 $\pm$ 1.18 & 64.00 $\pm$ 1.34 & 55.65 $\pm$ 4.60 & \textbf{78.67 $\pm$ 1.06} & 60.57 $\pm$ 1.15 & 85.00 $\pm$0.42 \\
Rwalk             & 55.91 $\pm$ 1.85 & 65.04 $\pm$ 0.11 & 53.66 $\pm$ 3.18 & 78.59 $\pm$ 1.37 & 59.03 $\pm$ 0.05 & \textbf{85.18 $\pm$ 0.57} \\
iCaRL             & 40.70 $\pm$ 5.13 & \textbf{65.61 $\pm$ 2.57} & 45.98 $\pm$ 3.04 & 57.07 $\pm$ 2.74 & 48.11 $\pm$ 4.63 & 64.90 $\pm$ 7.95 \\
GDumb             & 50.37 $\pm$ 1.17 & 42.47 $\pm$ 1.15 & 46.70 $\pm$ 1.53 & 43.16 $\pm$ 0.77 & 47.78 $\pm$ 3.77 & 45.72 $\pm$ 0.64 \\
BiC               & 44.70 $\pm$ 2.12 & 59.53 $\pm$ 4.30 & 42.06 $\pm$ 2.41 & 61.45 $\pm$ 6.25 & 42.92 $\pm$ 1.47 & 71.93 $\pm$ 2.45 \\
\cmidrule(lr){1-7} 
\textbf{\memname w/o DA}  & 54.05 $\pm$ 4.94 & 59.47 $\pm$ 0.61 & 60.87 $\pm$ 0.88 & 74.58 $\pm$ 0.60 & 60.92 $\pm$ 6.48 & 83.91 $\pm$ 0.40 \\ 
\textbf{\our}     & \textbf{66.25 $\pm$ 0.21} & 61.91 $\pm$ 0.63 & \textbf{71.13 $\pm$ 0.18} & 76.86 $\pm$ 0.04 & \textbf{73.90 $\pm$ 0.80} & 85.10 $\pm$ 0.16 \\
\bottomrule
\end{tabular}%
}
\vspace{-1em}
\end{table*}

\paragraph{On Various Blurry Levels. }
Even though blurry-CIL is the main task of our interests, it is interesting to investigate the performance of the proposed \memname on disjoint-CIL (\ie, Blurry0) setup and in various blurry levels.
We summarize the comparative results in \tablename~\ref{tab:cifar10-blurry-disjoint}.

In disjoint-CIL where catastrophic forgetting is more severe than blurry-CIL, regularization-based methods such as EWC~\cite{ewc} and Rwalk~\cite{rwalk} show competitive performances. 
It is expected that disjoint-CIL setup tends to exaggerate catastrophic forgetting that regularization-based methods aim to address (Section~\ref{sec:problem_def}).
Notably, \memname performs comparably  without any regularization while outperforming rehearsal-based methods, \eg, iCaRL, GDumb and BiC. 

In the offline setup, the gain by \memname diminishes and prior arts slightly outperform the \memname.
We conjecture that keeping the large incoming samples in buffer dilutes the sensitivity of exemplar sampling.
In blurry-CIL setups with online-setting (Blurry10 and Blurry30), \memname outperforms other baselines by remarkable margins even when DA is not applied.  
With the proposed DA, \our achieves over $70\%$ accuracy for both Blurry10 and Blurry30 setups, far better than the other baselines. 

We further compare the accuracy trajectories over the task streams; three streams generated from stochastically assigned function, $\psi(c)$, with different random seeds, for CIFAR10 and single stream for ImageNet and summarize the results in \figurename~{\ref{fig:allgraphs}}. 
For the online settings ((a), (c) and (d)), our \our consistently outperforms the other baselines over entire task stream.
However, in offline setting (b), \our comparably performs to the prior arts over the entire task stream as summarized in \tablename~\ref{tab:cifar10-blurry-disjoint}.

\begin{table*}[tp]
\centering
\caption{Comparison of last accuracy (A5 ($\uparrow$), \%) over methods with data augmentations in CIFAR10-Blurry10-Online. The results on $K=1,000$ is reported in the supplementary material. `CM+AA' refers to CutMix+AutoAug.}

\label{tab:abl-data-augmentation}
\resizebox{1.0\linewidth}{!}{%
\begin{tabular}{@{}crrrrrrrrrr@{}}
\toprule
& \multicolumn{5}{c}{K=200} & \multicolumn{5}{c}{ K=500} \\

Methods & None & CutMix & RandAug & AutoAug & \begin{tabular}[c]{@{}l@{}}CM+AA\end{tabular} & None & CutMix & RandAug & AutoAug & \begin{tabular}[c]{@{}l@{}}CM+AA\end{tabular} \\

\cmidrule(lr){1-1} \cmidrule(lr){2-6} \cmidrule(lr){7-11} 
EWC   & 40.0$\pm$2.1 & 41.9$\pm$1.0 & 44.7$\pm$0.6 & 48.3$\pm$3.5 & 50.3$\pm$1.2 
& 55.6$\pm$4.6 & 56.2$\pm$0.7 & 60.0$\pm$5.3 & 64.8$\pm$0.6 & 67.5$\pm$0.9 \\
Rwalk & 38.6$\pm$1.5 & 41.3$\pm$2.2 & 46.5$\pm$2.9 & 48.7$\pm$2.7 & 51.8$\pm$1.6
& 53.6$\pm$3.1 & 57.5$\pm$1.4 & 62.5$\pm$3.0 & 64.7$\pm$1.0 & 67.2$\pm$1.5 \\
iCaRL & 37.4$\pm$1.3 & 37.9$\pm$3.8 & 38.4$\pm$1.4 & 41.8$\pm$2.3 & 43.3$\pm$2.2
& 45.9$\pm$3.0 & 46.9$\pm$1.4 & 51.3$\pm$1.1 & 51.6$\pm$2.8 & 56.6$\pm$1.2 \\
GDumb & 33.3$\pm$2.0 & 35.8$\pm$1.0 & 37.1$\pm$2.0 & 38.4$\pm$1.1 & 41.4$\pm$1.1
& 46.7$\pm$1.5 & 49.4$\pm$1.0 & 54.3$\pm$1.4 & 55.9$\pm$1.4 & 58.2$\pm$2.7 \\
BiC   & 33.2$\pm$0.8 & 33.2$\pm$0.8 & 27.1$\pm$2.7 & 29.7$\pm$3.1 & 31.2$\pm$0.7
& 42.0$\pm$2.4 & 42.0$\pm$2.4 & 38.6$\pm$2.8 & 38.7$\pm$1.5 & 38.4$\pm$2.5 \\
\cmidrule(lr){1-11} 
\textbf{\our} & \textbf{44.4$\pm$1.4} & \textbf{45.9$\pm$2.4} & \textbf{49.9$\pm$2.9} & \textbf{55.3$\pm$2.2} & \textbf{54.6$\pm$1.6}
& \textbf{60.8$\pm$0.8} & \textbf{62.0$\pm$3.5} & \textbf{68.6$\pm$0.7} & \textbf{69.6$\pm$2.9} & \textbf{71.1$\pm$0.1} \\ \bottomrule
\end{tabular}%
}
\vspace{-1.0em}
\end{table*}

\begin{table}[h]
\centering
\caption{Comparison of uncertainty measures for \memname on CIFAR10-Blurry10-Online (K=500).}
\label{tab:unceratinty_measure}
\resizebox{0.8\linewidth}{!}{%
\begin{tabular}{@{}crrr@{}}
\toprule
       & No\_MC & RandAug\_MC & AutoAug\_MC \\ \midrule
A5 (\%) & 58.59  & 61.27       & 60.1  \\ \bottomrule
\end{tabular}%
}
\vspace{-0.5em}
\end{table}

\vspace{-1em}\paragraph{Uncertainty Measure.}
We compare three methods for estimating uncertainty by various Monte-Calro methods; (1) no MC (No\_MC), (2) RandAug-based (RandAug\_MC), and (3) AutoAug~\cite{AutoAugment}-based methods (AutoAug\_MC), summarizing the A5 results in \tablename~\ref{tab:unceratinty_measure}.
Note that RandAug\_MC and AutoAug\_MC also leverage configuring the stochastic data perturbation set as well as DA during training. 

As shown in the table, the two automated DA-based methods improve the accuracy compared to the No\_MC case, caused by diversity-enhanced configuration.
For measuring the uncertainty in our \memname, we use RandAug\_MC. 
\begin{table}[h]
\centering
\caption{Comparison of last accuracy (A5 ($\uparrow$), \%) over memory update methods without data augmentations in CIFAR10-Blurry10-Online. `CM+AA' refers to CutMix+AutoAug.}
\label{tab:memory_update_rule}
\resizebox{\linewidth}{!}{%
\begin{tabular}{@{}crrrrrr@{}}
\toprule
& \multicolumn{3}{c}{K=200} 
& \multicolumn{3}{c}{K=1,000} \\ 
Methods & None & CutMix & CM+AA
& None & CutMix & CM+AA \\ 
\cmidrule(lr){1-1} \cmidrule(lr){2-4} \cmidrule(lr){5-7}
Random & 24.1 $\pm$ 1.4 & 24.0 $\pm$ 1.0 & 22.4 $\pm$ 0.8  
& 46.7 $\pm$ 2.5 & 52.5 $\pm$ 4.2 & 52.7 $\pm$ 2.8\\
Reservoir & 38.0 $\pm$ 2.2 & 39.1 $\pm$ 0.8 & 49.4 $\pm$ 1.8 
& 64.6 $\pm$ 4.2 & 67.2 $\pm$ 5.3 & 75.5 $\pm$ 0.0\\
Prototype & 34.6 $\pm$ 0.5 & 33.8 $\pm$ 1.9 & 26.5 $\pm$ 3.9
& 48.1 $\pm$ 5.7 & 41.1 $\pm$ 4.1 & 29.3 $\pm$ 1.5 \\
\cmidrule(lr){1-7} 
\textbf{Uncertainty} (\memname) & \textbf{43.8 $\pm$ 1.2}
& \textbf{42.4 $\pm$ 1.8} 
& \textbf{52.2 $\pm$ 1.3} 
& \textbf{64.7 $\pm$ 4.1} & \textbf{71.8 $\pm$ 4.3} & \textbf{76.1 $\pm$ 1.1}\\ \bottomrule

\end{tabular}%
}
\vspace{-1.0em}
\end{table}

\vspace{-1em}\paragraph{Comparison to Other Memory Update Algorithms. }
\label{subsec:RMU}
To investigate exclusive gains by the memory update algorithm, we compare \memname with other memory update schemes while leaving other components unchanged and summarize results in \tablename~\ref{tab:memory_update_rule}.
The other algorithms include \textit{Random}, \textit{Reservoir}~\cite{reservoir} and \textit{Prototype}~\cite{icarl}.
Random selects new exemplars for the next episodic memory randomly from current exemplars and incoming samples. 
Reservoir conducts uniform random sampling on a unknown length task stream.
The prototype selects the samples where the extracted features are close to the feature mean of its own class.
As shown in the table, \memname outperforms all the augmentation conditions with different settings of $K$.

\vspace{-2em}\paragraph{Data Augmentation. }
We also investigate the effects of various DA methods on performances by comparing the adopted DA methods with others while other components unchanged in \tablename~\ref{tab:abl-data-augmentation}. 
As shown in the table, other methods also enjoyed the performance enhancement by DA same as \our.
However, the enhancement from CutMix + AutoAug used for \our is the most effective among all DAs.  
Note that even when adding various DA, \memname achieves the best performance surpassing the other baselines. 

\section{Conclusion}
We address a realistic and real-world class incremental (continual) learning setup where tasks share the classes, denoted as blurry-CIL.
To effectively address such scenario, we propose to enhance diversity of samples in an episodic (or representative) memory. 
Specifically, we propose a new diversity-enhanced sampling method using per-sample perturbation-based uncertainty.
In addition, we employ diverse sets of data augmentation techniques to further improve the diversity, that is representativeness and discriminativeness of exemplars, induced from the proposed memory update. 

In blurry-CIL scenarios on CIFAR10, CIFAR100, and ImageNet, our diversity-enhancing method (named \fullour or \memname) not only outperforms the state-of-the-art methods by large margins but also presents comparable performances on disjoint and offline CIL setups. 
We further investigate the effectiveness of the proposed method in various blurry setups and even in the disjoint setup, along with in-depth analysis for each proposed components.
As a future work, we will investigate the relationships between uncertainty-based memory update and data augmentation in training time and their effects on diverse CIL tasks.



\vspace{-0.5em}
{
\begin{singlespacing}
\footnotesize
\paragraph{Acknowledgement.}
JC is partly supported by the National Research Foundation of Korea (NRF) (No.2019R1C1C1009283) and Institute of Information \& communications Technology Planning \& Evaluation (IITP) grant funded by the Korea government (MSIT) (No.2019-0-01842, Artificial Intelligence Graduate School Program (GIST) and No.2019-0-01351, Development of Ultra Low-Power Mobile Deep Learning Semiconductor With Compression/Decompression of Activation/Kernel Data), and Center for Applied Research in Artificial Intelligence (CARAI) grant funded by Defense Acquisition Program Administration (DAPA) and Agency for Defense Development (ADD) (UD190031RD).
All authors thank Sungmin Cha (NAVER AI Lab) and Hyeonseo Koh (GIST) for discussions, and NAVER Smart Machine Learning (NSML)~\cite{kim2018nsml} team for GPU support.
\end{singlespacing}
}

{\small

\bibliographystyle{ieee_fullname}
\bibliography{main}
}

\newpage
\appendix

\twocolumn[{\centering{\textbf{{\large Supplementary Material for}\\ \vspace{0.5em} {\fontsize{15}{50}\selectfont Rainbow Memory: Continual Learning with a Memory of Diverse Samples} \vspace{1.0em} }}}]


\section{Accuracy Over the Tasks in Various CIL Setups}

\begin{figure*}[t]
    \centering
    \begin{subfigure}{0.33\textwidth}
        \includegraphics[width=0.99\columnwidth]{figures/resized_online_disjoint.pdf}
        \caption{CIFAR10-Disjoint-Online}
        \label{fig:online_disjoint_supp}
    \end{subfigure}
    \begin{subfigure}{0.33\textwidth}
        \includegraphics[width=0.99\columnwidth]{figures/resized_online_blurry.pdf}
        \caption{CIFAR10-Blurry10-Online}
        \label{fig:online_blurry_supp}
    \end{subfigure}
    \begin{subfigure}{0.33\textwidth}
        \includegraphics[width=0.99\columnwidth]{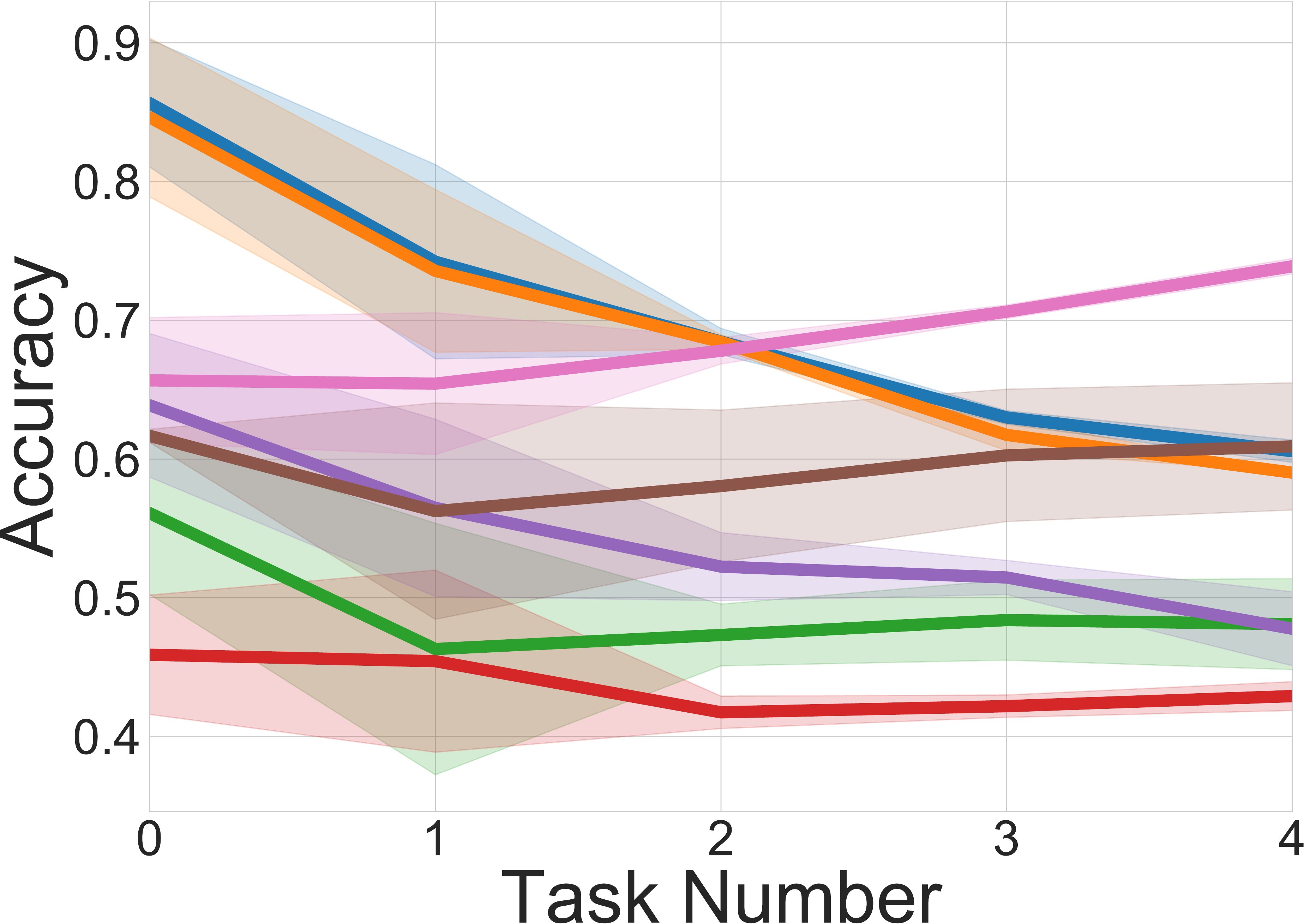}
        \caption{CIFAR10-Blurry30-Online}
        \label{fig:online_blurry30_supp}
    \end{subfigure}
    
    \caption{Illustration of accuracy changes as tasks are being learned in
    (a) CIFAR10-Disjoint-Online, 
    (b) CIFAR10-Blurry10-Online,
    (c) CIFAR10-Blurry30-Online settings.}
    \label{fig:online_settings_graphs}
\vskip -0.1in
\end{figure*}
\begin{figure*}[t]
    \centering
    \begin{subfigure}{0.33\textwidth}
        \includegraphics[width=0.99\columnwidth]{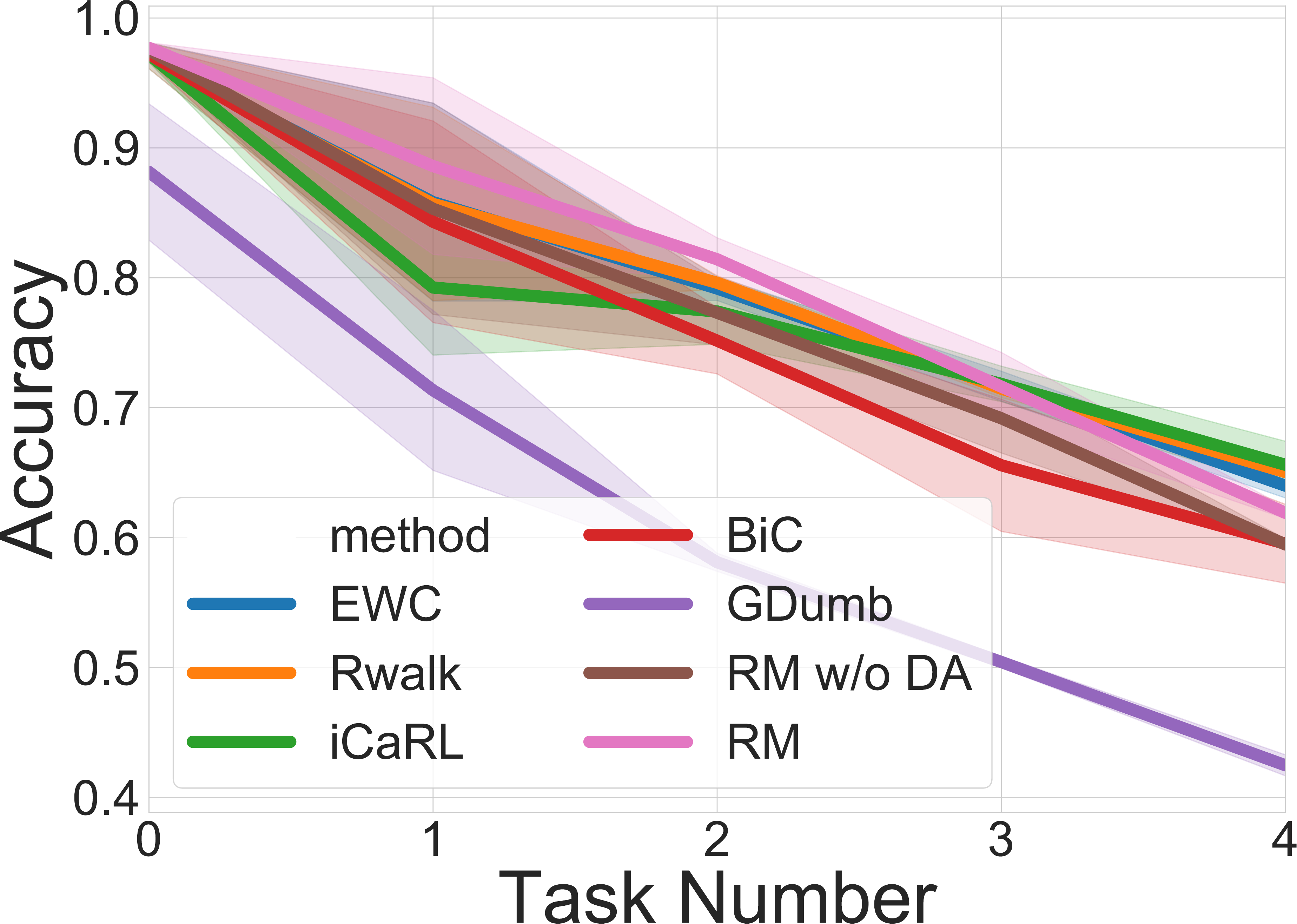}
        \caption{CIFAR10-Disjoint-Offline}
        \label{fig:offline_disjoint_supp}
    \end{subfigure}
    \begin{subfigure}{0.33\textwidth}
        \includegraphics[width=0.99\columnwidth]{figures/resized_offline_blurry.pdf}
        \caption{CIFAR10-Blurry10-Offline}
        \label{fig:offline_blurry_supp}
    \end{subfigure}
    \begin{subfigure}{0.33\textwidth}
        \includegraphics[width=0.99\columnwidth]{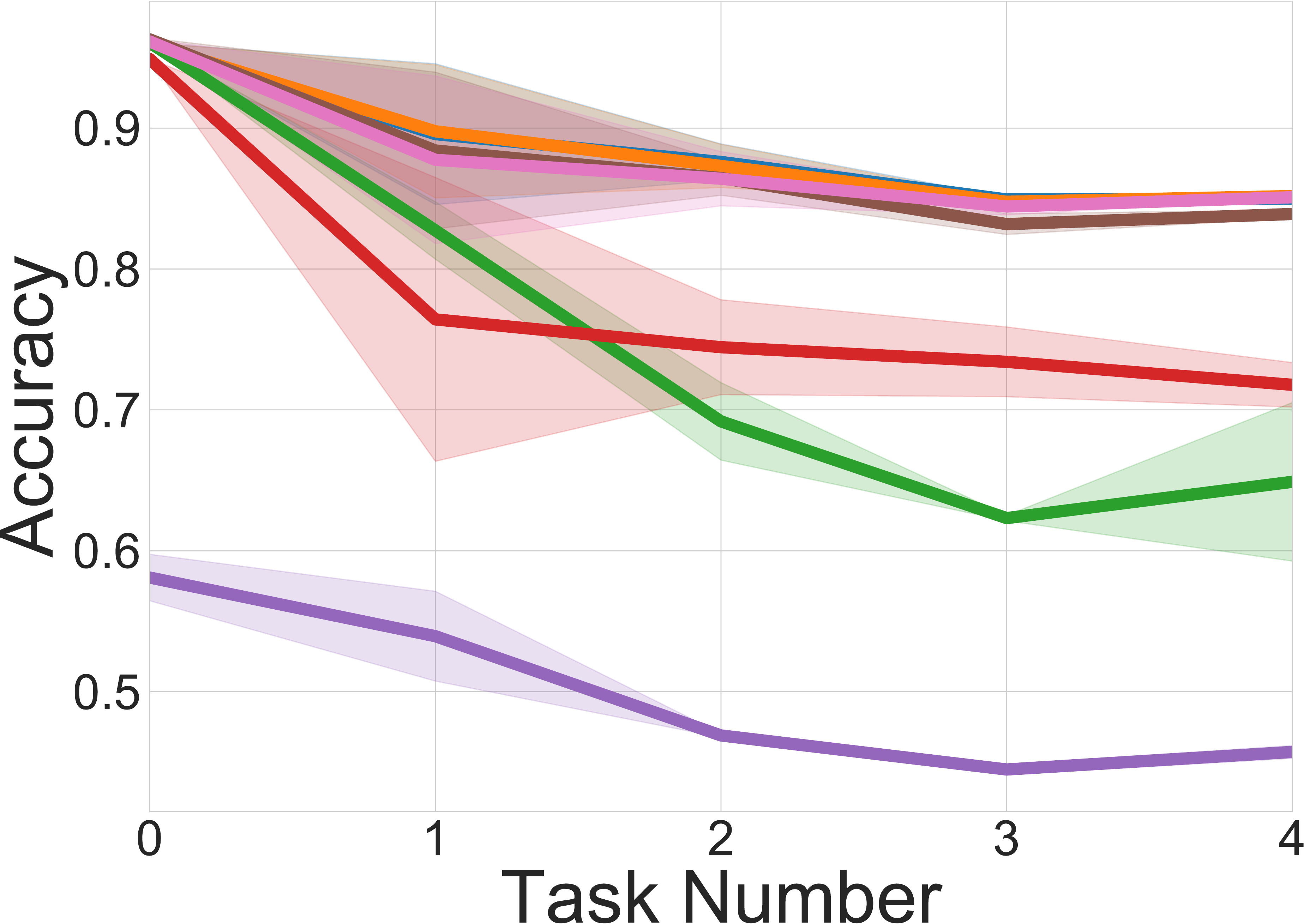}
        \caption{CIFAR10-Blurry30-Offline}
        \label{fig:offline_blurry30_supp}
    \end{subfigure}
    
    \caption{Illustration of accuracy changes as tasks are being learned in
    (a) CIFAR10-Disjoint-Offline, 
    (b) CIFAR10-Blurry10-Offline,
    (c) CIFAR10-Blurry30-Offline settings.}
    \label{fig:offline_settings_graphs}
\vskip -0.1in
\end{figure*}

The evaluation set for CIL methods consists of only the seen classes.
In disjoint setting, the number of seen classes increases when new tasks come, since classes of each task should be exclusive.
Therefore, classes of evaluation sets increase as the task iterations proceed and the accuracy tends to decrease (see \figurename~\ref{fig:online_disjoint_supp} and \ref{fig:offline_disjoint_supp}).

In blurry setting, on the other hand, the evaluation set comprises of entire classes as the tasks are not disjoint. 
Therefore, the model will see more data for each class as task iterations proceed; \eg, \figurename~\ref{fig:online_blurry_supp} and \ref{fig:offline_blurry_supp} show the accuracy increases in later tasks in Blurry10 configuration. 
Interestingly, as the blurry ratio increases (\eg, from Blurry10 to Blurry30), the accuracy flattens for all tasks as shown in \figurename~\ref{fig:online_blurry30_supp} and \ref{fig:offline_blurry30_supp}. 
We believe it is because the class frequency between minor and major classes in Blurry30 has less gap compared to Blurry10 so that the model can train well for all classes. Note that each task in the Blurry$M$ contains samples from its assigned major classes consisting of $(100-M)\%$ and ones of minor classes consisting of remaining $M\%$.


\figurename~\ref{fig:online_settings_graphs} and \ref{fig:offline_settings_graphs} show that our proposed approaches (RM w/o DA and RM) outperform other methods in the online setting, but the margin reduces or goes to negative in the offline setting as we mentioned in Section \textit{4.2 Results} in the main paper. 
It is because blurry-online setting allows to see the sample of current task once, and reuse only the exemplars stored in the memory. 
Hence, managing diversity in the memory is more crucial compared to offline setting, and thus maximally exhibiting the efficacy of our approaches.

\section{Metrics Details}
We use three metrics in Section \textit{4. Experiments} of the main paper; \textit{Last accuracy (A)}, \textit{Last forgetting (F)}, and \textit{Intransigence (I)} defined in~\cite{rwalk}.
Here, we describe them in detail.

\vspace{-1em}\paragraph{Last accuracy (A).}
Last accuracy reports an accuracy after entire training ends, thus it evaluates model over all classes being exposed during training. 

\vspace{-1em}\paragraph{Last forgetting (F).}
Forgetting measures how much the accuracy for each task is degraded (\ie, forgotten) compared to the best one in the training phases of previous tasks.
Hence, last forgetting reports an averaged forgetting metrics over all tasks after entire training ends.

\vspace{-1em}\paragraph{Intransigence (I).}
Intransigence measures the how much the accuracy for each task is achieved compared to the upper-bound, which comes from the non-CIL setting, then reports the average value for all tasks.
Therefore, as model learns new knowledge, intransigence will be improved.

\section{Class Distribution over Tasks}
As we mentioned in Section \textit{4.1 Experimental Setup} of the main paper, classes of CIFAR10 and CIFAR100 were randomly split into five tasks (2 and 20 classes per task, respectively), and classes of ImageNet were split into ten tasks to generate CIL-benchmark.
Moreover, we iterated every experiments three times with different class splits from three different random seeds except for ImageNet. 
Here, we summarize the class splits of CIFAR10 CIL-benchmarks used for our experiments in \tablename~\ref{tab:cifar10-class-split}.
We will release the splits and other configuration along with the code in our github repo: {\small\url{https://github.com/clovaai/rainbow-memory}}.

\begin{table}[]
\centering
\caption{Class splits for CIFAR10 CIL-benchmarks.}
\label{tab:cifar10-class-split}
\resizebox{0.98\linewidth}{!}{
\begin{tabular}{@{}llll@{}}
\toprule
 & Seed 1 & Seed 2 & Seed 3 \\ \midrule
Task 1 & truck, automobile & airplane, dog & ship, airplane \\
Task 2 & frog, airplane & ship, cat & dog, truck \\
Task 3 & cat, bird & horse, truck & automobile, frog \\
Task 4 & dog, horse & bird, frog & horse, cat \\
Task 5 & deer, ship & automobile, deer & bird, deer \\ \bottomrule
\end{tabular}
}
\end{table}

\section{Data Augmentation ($K=1,000$)}
As we mentioned in \textit{Table 4} of the main paper, we present the accuracy over methods with data augmentations in CIFAR10-Blurry10-Online when $K=1,000$ in \tablename~\ref{tab:abl-data-augmentation}. 
As shown in the table, it has the same tendency to the \emph{Table 4} of the main paper when $K$ is equal to 200 and 500. 
\memname performs the best with 78.0\%. 

\begin{table}[h]
\centering
\caption{Comparison of last accuracy (A5 ($\uparrow$), \%) over methods with data augmentations in CIFAR10-Blurry10-Online on $K=1,000$.}

\label{tab:abl-data-augmentation}
\resizebox{1.0\linewidth}{!}{
\begin{tabular}{@{}crrrrr@{}}
\toprule
Methods & None & CutMix & RandAug & AutoAug & \begin{tabular}[c]{@{}l@{}}CutMix\\ +AutoAug\end{tabular}  \\

\cmidrule(lr){1-1} \cmidrule(lr){2-6}

EWC  & 68.6$\pm$0.9 & 70.5$\pm$0.6 & 73.0$\pm$0.5 & 75.1$\pm$2.2 & 75.2$\pm$0.0 \\
Rwalk & 68.2$\pm$1.8 & 69.7$\pm$1.0 & 73.5$\pm$0.1 & 76.0$\pm$4.0 & 76.2$\pm$0.4 \\
iCaRL & 53.6$\pm$2.8 & 56.1$\pm$2.6 & 57.7$\pm$0.7 & 62.5$\pm$6.1 & 63.8$\pm$1.1 \\
GDumb & 59.1$\pm$0.3 & 64.2$\pm$1.2 & 67.5$\pm$1.3 & 67.6$\pm$2.2 & 70.3$\pm$0.6 \\
BiC   & 47.8$\pm$3.0 & 47.8$\pm$3.0 & 45.3$\pm$7.7 & 45.6$\pm$5.8 & 48.5$\pm$5.0 \\
\cmidrule(lr){1-6} 
\memname (Ours) & \textbf{70.9$\pm$1.5} & \textbf{74.7$\pm$0.7} & \textbf{76.4$\pm$0.4} & \textbf{77.5$\pm$0.7} & \textbf{78.0$\pm$0.5} \\ \bottomrule
\end{tabular}%
}
\end{table}

\end{document}